\documentclass[letterpaper, 10 pt, conference]{ieeeconf}  

\IEEEoverridecommandlockouts  

\usepackage{graphicx}
\usepackage{kantlipsum}
\usepackage[utf8]{inputenc} 
\usepackage[T1]{fontenc}    
\usepackage{hyperref}       
\usepackage{url}            
\usepackage{booktabs}       
\usepackage{amsfonts}       
\usepackage{nicefrac}       
\usepackage{microtype}      
\usepackage{graphicx, multicol} 
\usepackage{multirow}
\usepackage{array}
\usepackage{subfigure}

\usepackage{wrapfig}
\newcolumntype{H}{>{\setbox0=\hbox\bgroup}c<{\egroup}@{}}
\usepackage[export]{adjustbox}
\usepackage[utf8]{inputenc}
\usepackage[english]{babel}
\usepackage{amsmath}

\usepackage{tablefootnote}
\usepackage{threeparttable}
\let\labelindent\relax
\usepackage{enumitem,kantlipsum}
\usepackage{float}
\usepackage{tabulary,booktabs}
\newcommand{\RomanNumeralCaps}[1]

\title{\LARGE \bf
A data-set of piercing needle through deformable objects\\ for Deep Learning from Demonstrations
}

\author{Hamidreza Hashempour$^{\dagger,2}$, Kiyanoush Nazari$^{\dagger, 1}$, Fangxun Zhong$^3$ and Amir Ghalamzan E. $^{\dagger, 1}$
\thanks{$^{\dagger}$These authors have equal contribution.}
    \thanks{$^1$University of Lincoln, Intelligent Manipulation Lab (IML)
            {\tt\small aghalamzanesfahani@lincoln.ac.uk}; $^2$ University of Tehran; $^3$ T Stone Robotics Institute and Dept. of Mechanical and Automation Engineering, The Chinese University of Hong Kong, HKSAR, China.}%
}

\begin{document}

\maketitle
\thispagestyle{empty}
\pagestyle{empty}


\begin{abstract}

 Many robotic tasks are still teleoperated since automating them is very time consuming and expensive. Robot Learning from Demonstrations (RLfD) can reduce programming time and cost. However, conventional RLfD approaches are not directly applicable to many robotic tasks, e.g. robotic suturing with minimally invasive robots, as they require a time-consuming process of designing features from visual information. Deep Neural Networks (DNNs) have emerged as useful tools for creating complex models capturing the relationship between high-dimensional observation space and low-level action/state space. Nonetheless, such approaches require a dataset suitable for training appropriate DNN models. This paper presents a dataset of inserting/piercing a needle with two arms of da Vinci Research Kit (DVRK) in/through soft tissues. The dataset consists of (i) 60 successful needle insertion trials with randomised desired exit points recorded by 6 high-resolution calibrated cameras, (ii) the corresponding robot’s data, calibration parameters and (iii) the commanded robot’s control input where all the collected data are \emph{synchronised}. The dataset is designed for Deep-RLfD approaches. We also implemented several deep RLfD architectures, including simple feed-forward CNNs and different Recurrent Convolutional Networks (RCNs). Our study indicates RCNs improve the prediction accuracy of the model despite that the baseline feed-forward CNNs successfully learns the relationship between the visual information and the next step control actions of the robot. The dataset, as well as our baseline implementations of RLfD, are publicly available for bench-marking\footnote{{\tt https://github.com/imanlab/d-lfd}}.

\end{abstract}

\section{INTRODUCTION}
\begin{figure}[tb!]%
\centering
\subfigure[][]{%
\label{fig:ex3-a}%
\includegraphics[height=2.6cm]{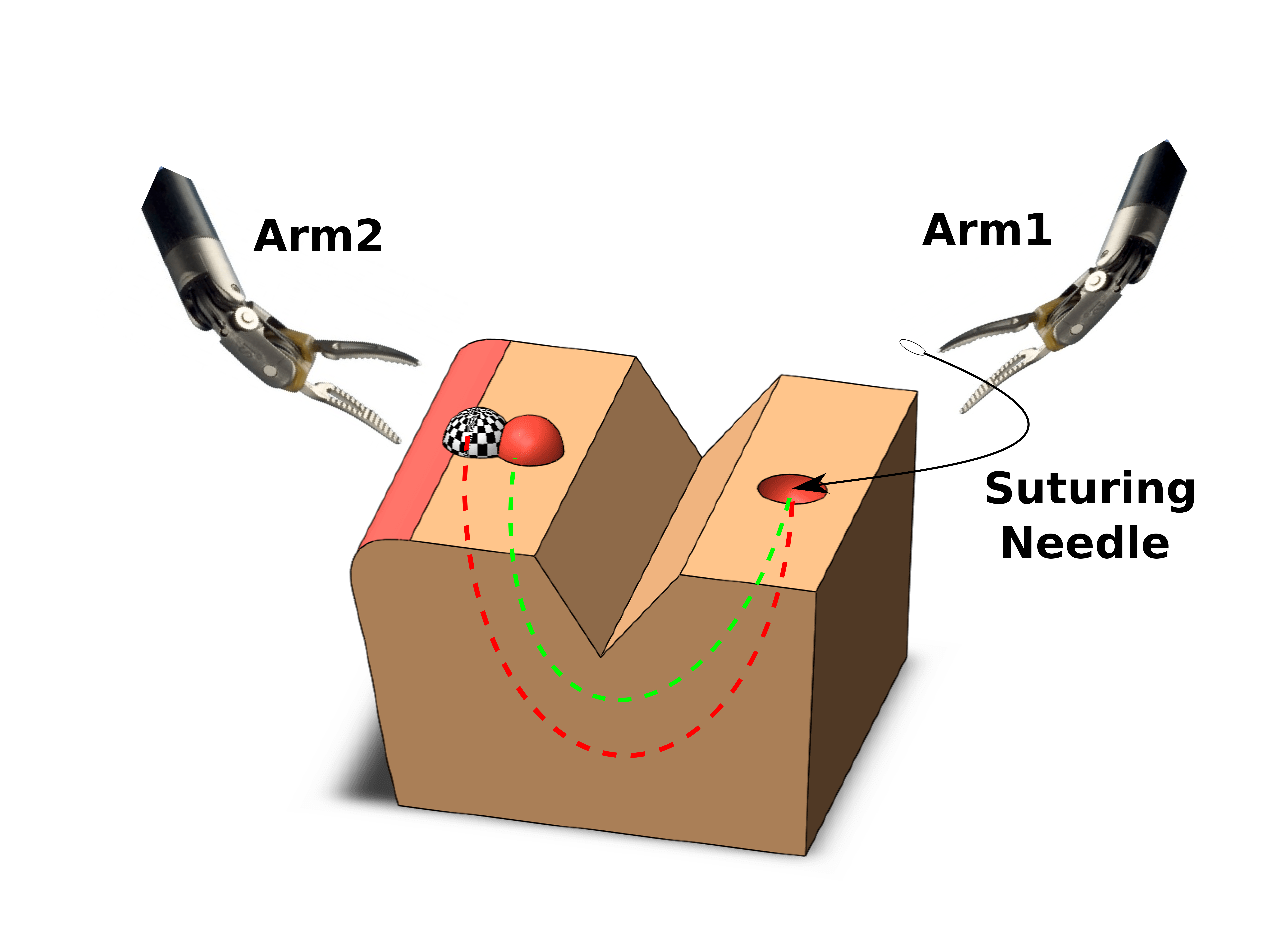}}%
\subfigure[][]{%
\label{fig:ex3-b}%
\includegraphics[height=2.4cm]{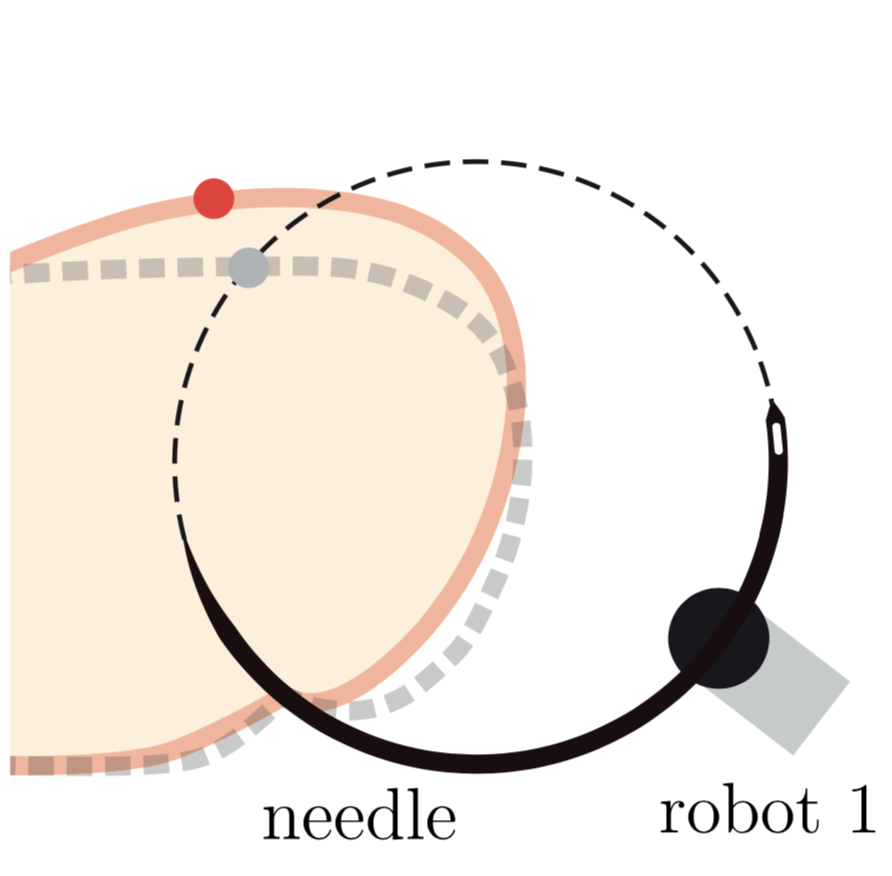}} 
\subfigure[][]{%
\label{fig:ex3-c}%
\includegraphics[height=2.4cm]{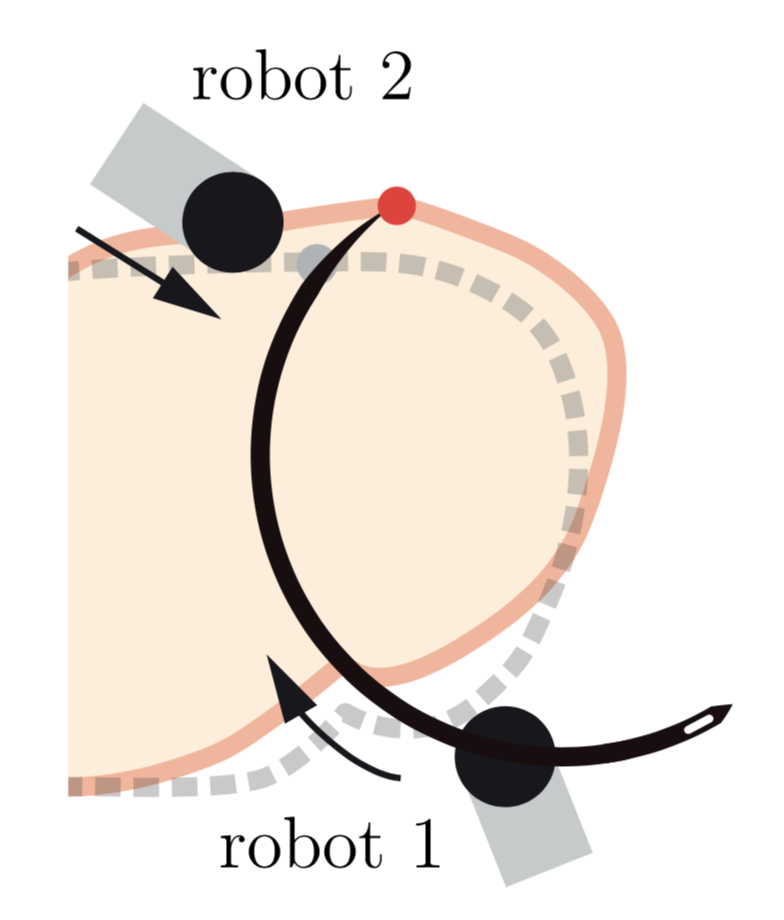}}%
\vspace{-8pt}
\caption{Needle insertion in soft tissue:
\subref{fig:ex3-a} the needle pushes the tissue making the actual (red sphere) and desired (black-white sphere) needle exit points as well as the actual (green dashed line) and desired (red dashed line) needle paths different;
\subref{fig:ex3-b} A 2D schematic of tissue deformation due to needle insertion (original shape shown as grey dash line) alter the target (red dot) from its original position (grey dot); and
\subref{fig:ex3-c} Arm 2 is actively manipulating the tissue to improve insertion accuracy, as adopted by manual suturing skills.}%
\vspace{-11pt}
\label{fig::tissue}%
\end{figure}

Robotic technology is increasingly considered as the mean of performing many tasks, especially in medical applications such as Minimally Invasive Surgery (MIS) \cite{guthart2000intuitive, mirbagheri2020sina} and orthopaedic surgery \cite{jacofsky2016robotics}. One of the common robotic tasks in MIS is suturing which usually takes a considerable amount of operation time. Suturing contributes to the increased fatigue and cognitive loading on the surgeon. Autonomous robotic suturing has been of noticeable interest \cite{moller2020laparoscopic} to shorten the operation time and decrease the cognitive load. Suturing requires bi-manual robotic operations, precise position control and a small margin of error which yields a minimal invasive effect on the patient. Suturing patients' tissues impose very interesting and challenging research questions in terms of robot perception, planning and control because of the interaction of the robot and a deformable tissue. During inserting the circular needle (see Fig.~\ref{fig::tissue}) into the deformable tissue, the needle tip pushes the tissue, hence, the desired and actual path differ and the actual exit-point, shown by the black-white sphere in Fig.~\ref{fig::tissue}), drifts away from the desired exit-point (red sphere) resulting in a less effective grip for the stitch or (in some cases) failure of the stitch- as the stitch cuts the tissue. 

\begin{figure}[bt!]
\vspace{-0pt}
\centering
\subfigure[][]{%
\label{fig:ds-setup}%
\includegraphics[height=5cm, width=7cm]{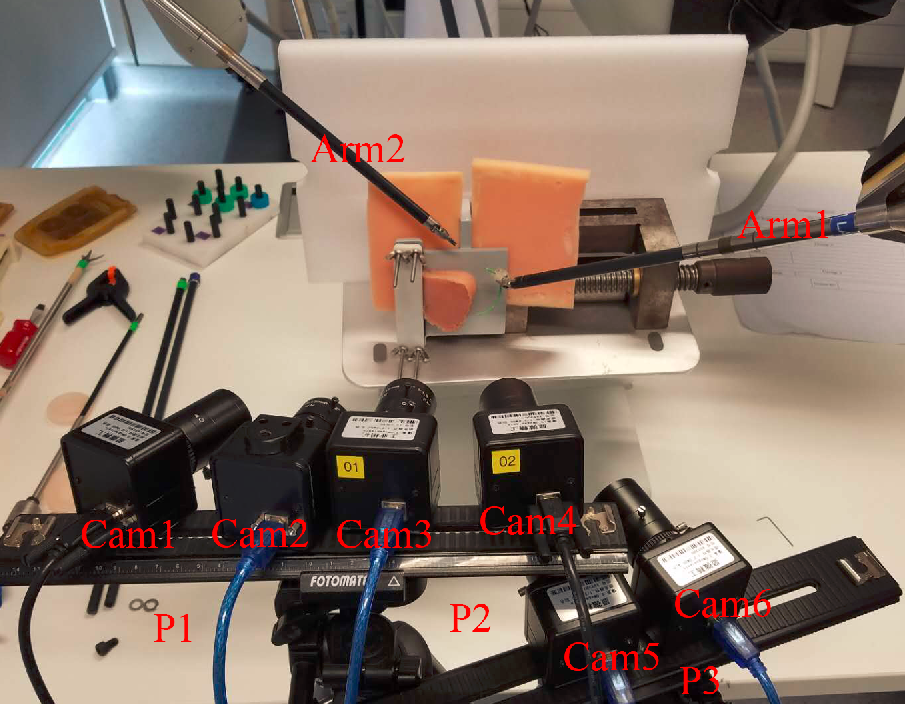}}
\hspace{1pt}%
\subfigure[][]{%
\label{fig:ds}%
\includegraphics[ height=1.25cm]{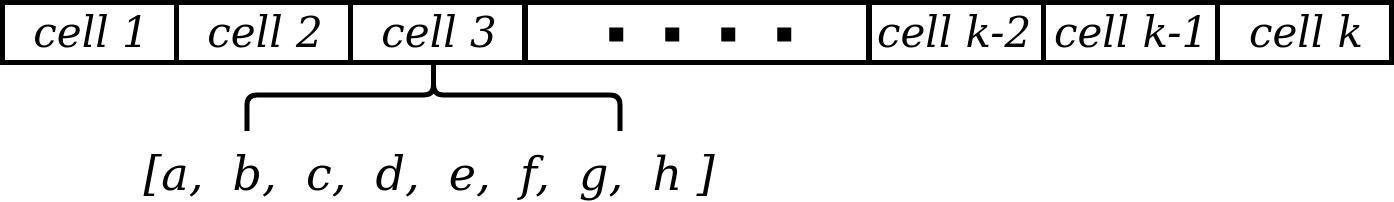}}
\vspace{-8pt}
\caption{\subref{fig:ds-setup} Dataset collection setup; \subref{fig:ds} dataset includes $a$ : joint space kinematic data of Arm 1 $(1{\times}6)$; $b$: joint space kinematic data of Arm 2 $(1{\times}6)$; $c$ : pose of Arm 1 w.r.t its base frame $(4{\times}4)$; $d$ : pose of Arm 2 w.r.t its base frame $(4{\times}4)$; $e$: pose of Arm 1 w.r.t its base frame at $t+1$; $f$: pose of Arm 2 w.r.t its base frame at $t+1$; $g$: 2D tracking target point on the image captured by 6 cameras $(6{\times}2)$; $h$: computed 3D position of the target point w.r.t Camera 3 $(1{\times}3)$}
\vspace{-.8 cm}
\label{fig:setup}
\end{figure}

In practice, surgeons utilise Arm 2  to manipulate the tissue and ensure the desired and actual exit-points are the same. Surgeons only use visual feedbacks to predict the needle exit point (performed by Arm 1 in Fig.~\ref{fig:setup} and \ref{fig::tissue}) and close the control loop by commanding Arm 2 to push/pool the tissue. 
Formulating such tasks and implementing the corresponding automatic controller is very complex and time-consuming.  
Classically, Robot Learning from Demonstrations (RLfD) has been proposed to reduce the programming time and cost by learning such a control policy from sample successful task executions which may be teleoperated, e.g. teleoperated needle insertion. 
In similar applications in medical robotics, various control strategies have been proposed for robotic hands position control \cite{ng2019precision} and autonomous robotic suturing \cite{hauser2009feedback}, \cite{sundaresan2019automated}. 
However, hand designed features extracting semantically meaningful information from high-dimensional images \cite{alterovitz2005planning,reed2011robot} limits the scalability/generalisability of such approaches. 
An end-to-end data driven approach which efficiently finds the mapping between the optimal control actions demonstrated in the dataset of successful completion of the task and high-dimensional image space as well as robot states can significantly helps automation of several complex tasks. 
To have precise control in such unstructured environments, we propose a framework of Deep-RLfD which allows to learn the controller of complex task such as the needle insertion.

The contribution of this paper is twofold: (i) we present a dataset of needle insertion into deformable objects. The needle insertion task is complex and involves highly nonlinear interactions between the robot and the tissue. 
As such, the framework allowing learning the controller to perform the task can be used for other complex task too. 
The dataset includes 60 successful needle insertion experiments. 
Dataset has several interesting features opening avenues to future studies in the context of Deep-RLfD: (1) the experiments are recorded by 3 pairs of stereo cameras calibrated w.r.t. the base frame of Arm 1 (in total 6 camera views as well as their calibration parameters w.r.t. to Arm 1 and Arm 2 are available.). This allows future studies on generalising the controller to different camera view. (2) Moreover, the camera calibration provide useful information about the camera views, hence, can be used by the model to improve the generalisation capability of the model; (3) the dataset includes all the robot data synched with the corresponding videos including the joint position and end-effector pose at each time step; 
\newline
(ii) We propose a novel Deep Robot Learning from Demonstrations (D-RLfD) framework and employ state-of-the-art Deep Neural Networks architectures in this framework. We illustrate the effectiveness of our D-RLfD framework in learning control actions (i.e. positions and orientations) for the robot hands to insert a needle in a deformable tissue. 
Using the state of the art DNN architectures, the developed D-RLfD yields high accuracy in predicting control actions based on the visual sensory information provided by a camera.

\section{Related Works}
Conventional RLfD~\cite{argall2009survey}, imitation learning~\cite{hussein2017imitation}, behavioural cloning~\cite{torabi2018behavioral} and other such approaches usually require low dimension feature describing robot workspace. For instance, the obstacle position and velocity and initial and goal point for planning and control the manipulative movements. In fact, such approaches may not be capable of tackling high dimensional visual information or video feed of the robot's workspace. Recent approaches in video prediction~\cite{mathieu2015deep} illustrated a step towards learning an end-to-end RLfD model. 

Deep neural networks (DNN) have been utilised for video prediction and state estimation based on visual sensory information~\cite{finn2016unsupervised, becker2019recurrent, haarnoja2016backprop}. In the context of RLfD, however, we need to build a model learning the robot controller based on available sample demonstrations. In conventional RLfD~\cite{argall2009survey}, it is highly desired to build a model which learns a task only from a few demonstrations facilitating Human-Robot Collaboration, i.e. a non-expert person can provide the robot with a few samples of task completion and the robot learns how to do it.  In contrast, we aim at solving the problem of RLfD where there are big datasets readily available, e.g. in robotic surgery all the operations data can be logged and may be readily available to train Deep RLfD.

In this work, we present a dataset that can be used for RLfD. Moreover, we present several deep neural networks architectures including convolutional and recurrent neural networks, as well as, the combination of them (RCN) to learn the robot control for the task of inserting a needle in a deformable tissue which is an important/challenging task in robotic surgery and suturing. This task is highly non-linear as it involves the interaction between the robot and the deformable object. Data-driven approaches are capable of modelling highly nonlinear dynamic systems if they are provided with enough training data. For sensitive applications such as robotic surgery, a system with highly nonlinear observation data \textit{e.g.} image sequence, the maximum value of tracking error plays a significant role in evaluating the reliability of the system. We present a discriminative model which minimizes the tracking error relative to similar approaches. In order to choose the most efficient model, various SOTA architectures are tested and by tuning the hyper-parameters of the network the achieved accuracy and error values are optimized.
\newline
%
%
\textbf{Control theory for robotic surgery.}
Classic and modern control theories have been employed to precisely control medical robots, e.g., \cite{saeidi2019autonomous} developed an autonomous laparoscopic robotic suturing system in which point cloud data is used to construct the tissue surface and plan the suturing points. In \cite{chow2014novel}, a vision-guided knot-tying method is proposed for autonomous robotic suturing. Needle path planning in autonomous robotic suturing is investigated in \cite{jackson2013needle}. \cite{shademan2016supervised} proposed a supervised autonomous robotic suturing system in which three dimensional infrared imaging is used for point cloud construction and then detecting tissue deformation. However, tissue was first marked by syringe with markers to track motions which may have invasive effects. Visual servoing has been used for suturing in micro-surgical task \cite{zong2006visually} and steering flexible needle \cite{chatelain20153d}. However, these approaches usually need time consuming hand-designed low dimensional features extracted from high dimensional visual information of the robot workspace. 
\newline
\textbf{Robot Learning from Demonstration (RLfD).}
\cite{atkeson1997learning} is one of the earliest works on Robot Learning from Demonstration (RLfD) that enabled further research in robotic surgery. For instance, M. Power \textit{et al.} \cite{power2015cooperative} used continuous HMMs to recognise the sub-tasks and learn needle passing and peg-transfer with Haptic feedback to enhance teleoperation. GP is used in Bi-manual knotting task with da Vinci robot~\cite{osa2014trajectory} to adapt the movement of the Arm 1 to the movements of the Arm 2 and the rope shape. However, these are only applied to low dimensional observations. Schulman \textit{et al.} \cite{schulman2013case} have used suturing trajectory transfer from human to a Raven robot based on 3D point cloud data. Nonetheless, the algorithm is not fully autonomous and requires a human's input for detecting the landmarks. Yang \textit{et al.}~\cite{yang2018learning} used DMPs in a superficial suturing task by decomposing the task into simpler sub-tasks. However, it is limited to a single class of superficial suturing and the error values are not reported in their work. In \cite{huang2016vision} RLfD is used in a vision-based swing system in the textile industry. While the motion of the hand which grips the needle is generated by an RLfD model, the other robotic hand's motion depends on the object on which sewing is performed. 
These approaches are limited to low dimensional observation and cannot be applied where only high-dimensional visual information (video or images) is available. Deep Neural Networks emerged as handy tools for dealing with high dimensional visual observations. \newline
\textbf{Time series forecasting/vision in robotic surgery by DNNs}
Wang \textit{et al.} \cite{wang2018deep} used deep CNNs for skill assessments in surgical robots by learning an expert deep model from successful task completions. Zhou \textit{et al.} \cite{zhou2019needle} proposed a DNN structure for sub-retinal injection and eye surgery with a high level of accuracy for needle detection. Extending this concept to learning a controller from a dataset is an interesting research question. \cite{marban2017estimating} uses CNN and LSTM in a network to estimate surgical tool's tip's position and velocity with video sequence input on JIGSAW dataset~\cite{gao2014jhu} which is a surgical activity dataset for human motion modeling for 3 elementary tasks including suturing, knot-tying and needle-passing teleoperated by eight expert subjects. The gestures defined in suturing in JIGSAW dataset does not include tissue manipulation and both robotic hands manipulate only the needle and the suture. \cite{becker2019recurrent, haarnoja2016backprop} used Kalman filter in a DNN for state estimation in base-line tasks only in simulation. \cite{watter2015embed} uses DNN with auto-encoders and MPC for robot control from raw images and is applied only on a toy dataset. However, neither of these works learns a robot controller from sample demonstrations. We present a dataset of inserting needle into a deformable tissue suitable for deep-RLfD approaches. The dataset includes videos captured by 6 cameras calibrated w.r.t. the robot base frame, along with the corresponding calibration parameters, robot states--position and orientation (pose) of the Arm 1 and Arm 2--at time $t$, and the robot control inputs (which are the desired pose of Arm 1 and Arm 2 at time $t+1$). 
We also present some baseline deep NN models which use a complex feature extractor on a real-world robotic image dataset to generate latent vector concatenated with robot state data and calibration parameters and fed into a recurrent neural network for deep time series prediction. 

\section{Needle Insertion Data-Set}
\vspace{-1pt}
We have created a dataset of inserting a needle into a deformable object using a da Vinci Research Kit (DVRK), see Fig.~\ref{fig:setup} for the experimental setup used for data collection. 
The phantoms (deformable objects used in our experiments) are made of homogeneous polyethylene whose Young's modulus is circa 0.02~0.04 [GPa]. Inserting the circular needle in (and piercing it thorough) a soft tissue (Fig.~\ref{fig:ex3-b}) deforms the tissue such that not enough tissue supports the stitch which may result in failure as the stitch may cut the tissue. In robotic surgery, Arm 2 is used to manipulate the soft tissue (Fig.~\ref{fig:ex3-c}) to ensure the needle pass/exits through the desired point yielding enough gripped tissue by the stitch. This procedure is delicate and requires significant training by novice surgeons and automating this by Deep-RLfD is of notable interest.

The experimental setup we used for data collection is shown in Fig.~\ref{fig:setup}. We use two Patient-Side Manipulators (PSMs) from the da Vinci Research Kit (namely Arm 1 and Arm 2 in Fig.~\ref{fig:setup}) to perform needle insertion in a soft and deformable object. The circular needle is fixed by a Large Needle Driver (LND) on Arm 1 and  \emph{ProGrasp Forcep} (PGF) on Arm 2 is utilised to grip the tissue and manipulate it.
Three pairs of fixed stereo cameras are used to record the video of the experiments in which the desired exit point of the needle on the tissue surface and the needle tip is marked and tracked. 
All the cameras are calibrated and the calibration parameters of the left cameras of each stereo camera w.r.t base frame of Arm 1 (${}^{a_1}T_{l_i}$ where $i$ denotes the camera number) and Arm 2 (${}^{a_2}T_{l_i}$) are included in the dataset. 
Moreover, the calibration parameters of the right camera in every stereo camera is also included in the dataset (${}^{r_i}T_{l_i}$). 
One can readily obtain the eye-to-hand transformations between the right cameras and Arm 1 or Arm 2 by multiplying two transformation matrix (${}^{a_1}T_{r_i} \:= \:{}^{a_1}T_{l_i} \: {}^{l_i}T_{r_i}$).  

\setlength{\intextsep}{5pt}
\begin{figure}[tb!]
\centering
\includegraphics[height=5cm, width=8cm]{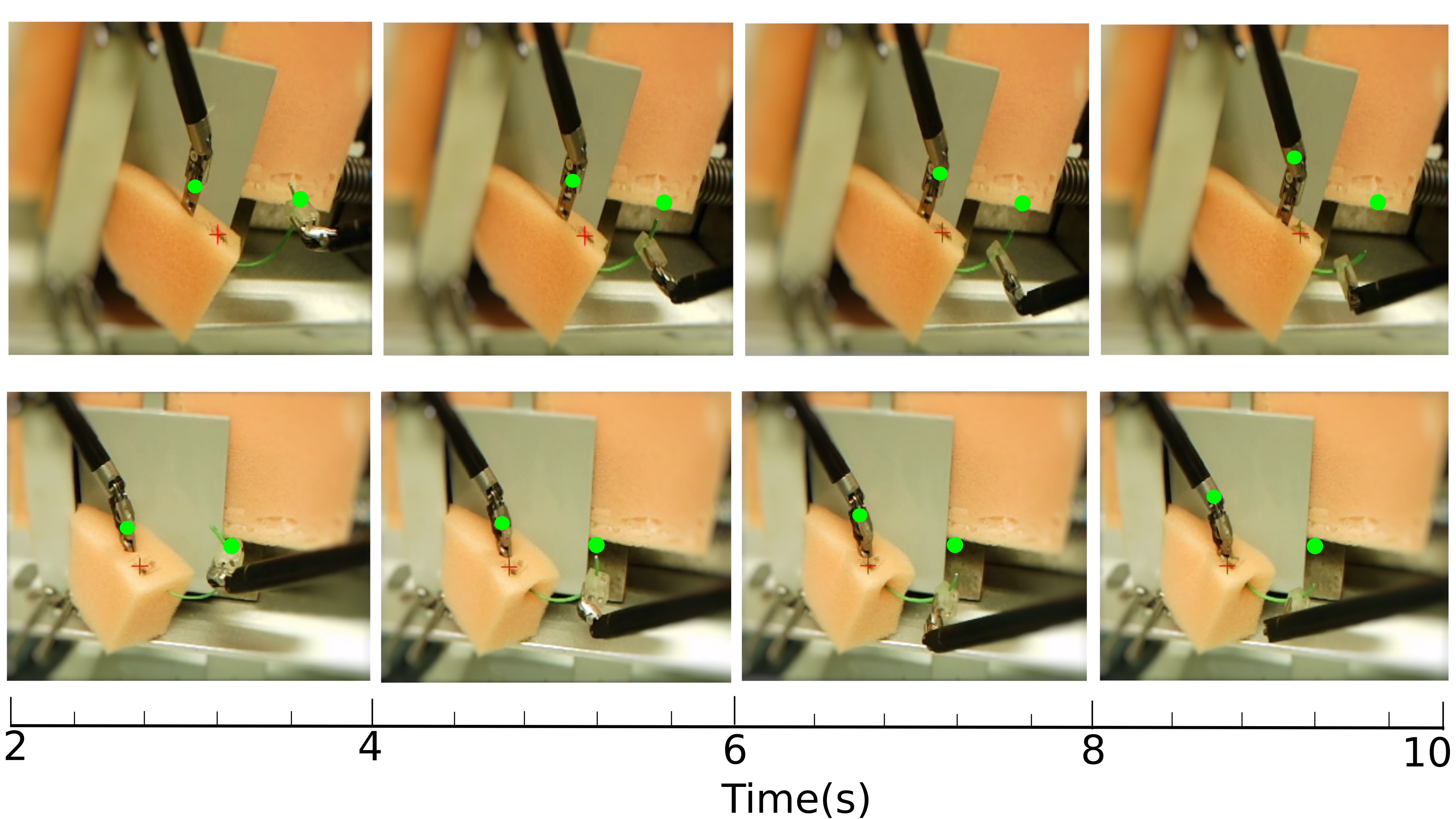}
\vspace{-5pt}
\caption{Sample image sequence from Camera 1 (top-row) and Camera 4 (bottom-row) viewpoints. The task is to track the red cross sign which exist in the videos. The green points show the initial position of the grippers in pixel space and are just added here to illustrate the motion of the hands.}
\label{fig:imgsequence}
\vspace{-10pt}
\end{figure}

The 3D pose of the surgical needle is online localised \cite{zhong2019dual} during the dual-arm robot movements using the extended forward kinematic. There is no information regarding mechanical properties of the tissue in the dataset. 
Each camera captures videos with 310x244 resolution at 25 fps.
The task averagely takes $\sim$170-200 control loops (costing $\sim$7-10s) for completion.
We performed 60 needle insertion and piercing through soft tissue with different tissue configurations and collected all the corresponding information, including (i) videos with 6 different camera views, (ii) the corresponding hand-to-eye calibration parameters, (iii) poses of Arm 1 and Arms 2 at time $t$ (iv) control commands--desired pose for Arm 2, (v) the tracking of the needle tip during piercing through the tissue, and (vi) the desired needle exit point on each video at each time $t$. 
The exit point at each trial is randomly selected on the tissue surface (tracked and visualised in the videos by a red cross mark on the image). The initial grasping configuration of Arm 2 on tissue also differs across different trials.

The dataset we collected consists of an array of cells at each sampling time step $t$ (each trial may have 140-200 cells depends on the task duration.) where each cell includes $a$ : kinematic data of Arm 1 in joint space $(1{\times}6)$; $b$: kinematic data of Arm 2 in joint space $(1{\times}6)$; $c$: pose ($SE(3)$ where $SE(3)$ denotes the  group of 3D poses\footnote{$SE(3)$  $=$ $R^3\times SO(3)$ and $SO(3)\subset R^{3\times3}$ denotes the group of rotations in three dimensions}--3D position $p$ and 3D orientation $r$) of Arm 1 w.r.t its base frame $(4{\times}4)$; $d$ : pose of Arm 2 w.r.t its base frame $(4{\times}4)$; $e$: pose of Arm 1 w.r.t its base frame at $t+1$; $f$: pose of Arm 2 w.r.t its base frame at $t+1$; $g$: 2D tracking target point on the image captured by 6 cameras $(6{\times}2)$; $h$: computed 3D position of the target point w.r.t Camera 3 $(1{\times}3)$.

Thus, the robot state is denoted as a transformation matrix:
\begin{equation}
S_t = 
\begin{bmatrix}
R_t & p_t \\
0 & 1
\end{bmatrix}
= 
\begin{bmatrix}
r_{11} & r_{12} & r_{13} & p_x\\
r_{21} & r_{22} & r_{23} & p_y\\
r_{31} & r_{32} & r_{33} & p_z \\
0 & 0 & 0 & 1
\end{bmatrix}
\end{equation}

Furthermore, the camera calibration data of the system is saved which contain: (i) eye-to-hand calibration results (the pose of left camera of the pair in stereo camera Px w.r.t. arm Rx), (ii) calibration parameters of three stereo camera pairs (P1: left pair, P2: middle pair, P3: right pair in Fig.~\ref{fig:ds-setup}).
We decided to use unit quaternions to represent poses which is robust to the singularity issue, a common issue with other choices of orientation representations.  

We preprocessed the dataset and created images from the video frames corresponding to the sampling time of the robot data collected during the experiments; the images are cropped and resized to $224\times224$. As  an illustration, figure \ref{fig:imgsequence} shows a sample sequence of four images captured with two seconds intervals from Camera 1 and Camera 4. The red cross sign in the videos shows the desired needle tip exit point which needs to be tracked. We also converted all the rotation matrices to quaternions. The state of the robot is represented by a 7 dimensional vector $S_t = [q_{0} ,\; q_{1} ,\; q_{2},\; q_{3} ,\; p_x ,\; p_y ,\; p_z] \;\in\; R^{1\times7}$ including quaternion and position.

\section{Model Description}
\begin{figure*}[tb!]
\centering
\vspace{-1pt}
\includegraphics[height=5cm, width=13cm]{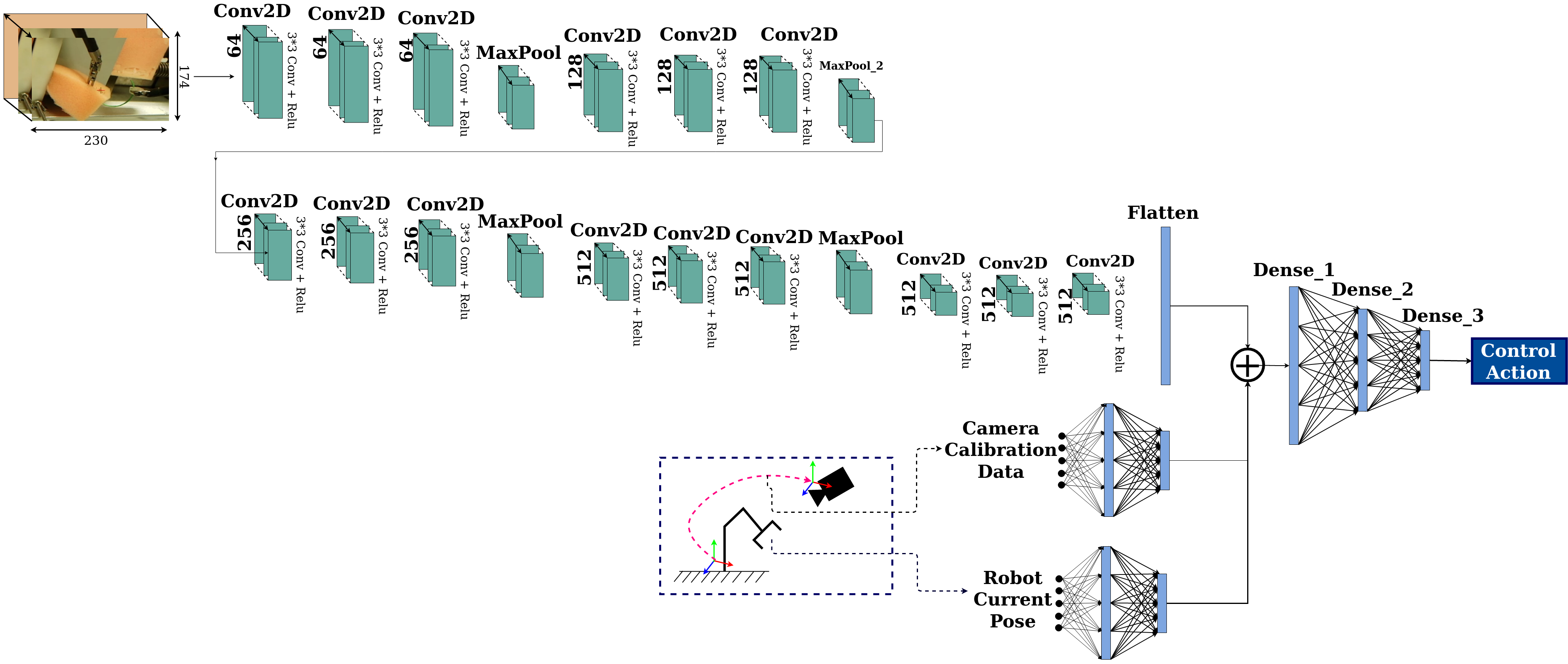}
\caption{Feed-forward CNN model}
\vspace{-10pt}
\label{fig:CNN}
\end{figure*}

\begin{figure*}[tb!]
\centering
\includegraphics[width=0.65\textwidth]{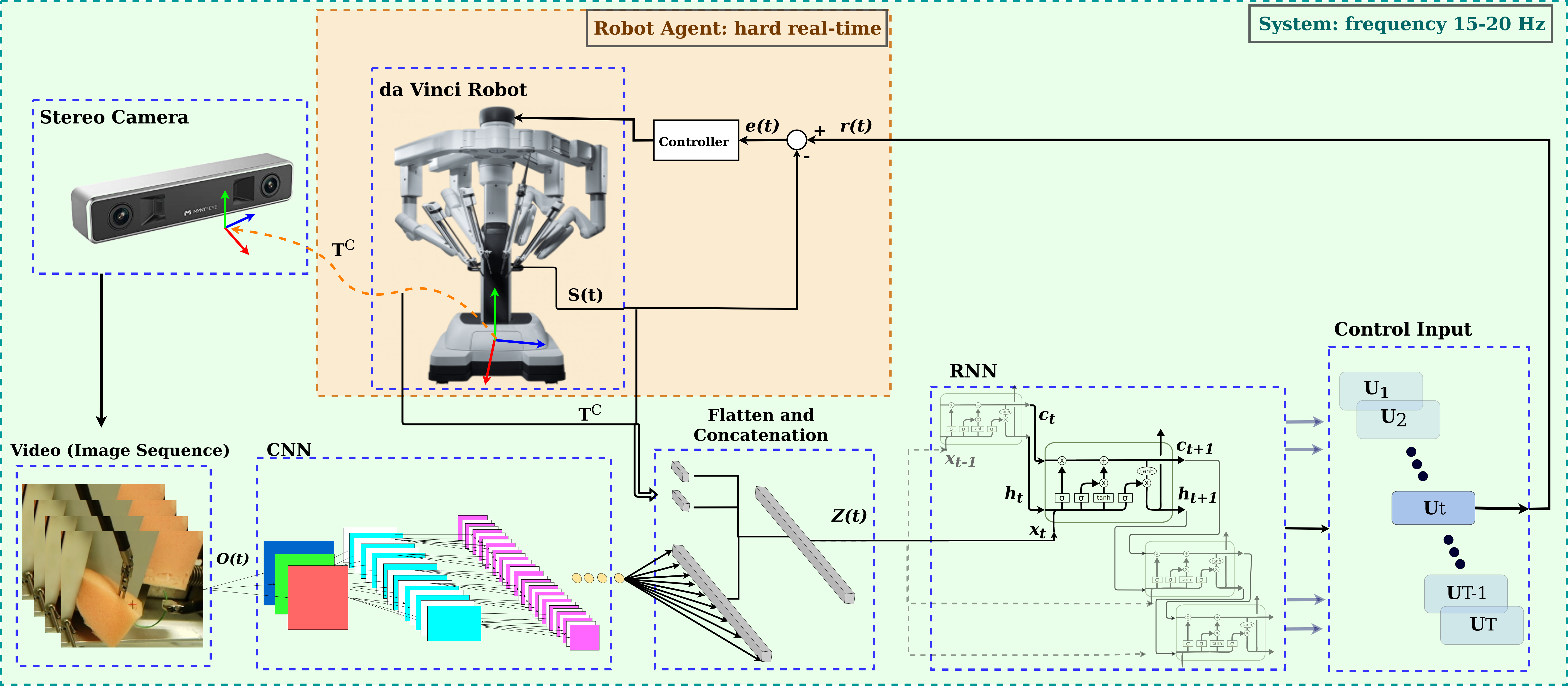}
\caption{System demonstration: Robot agent, Camera, and the network including CNN, concatenation and RNN blocks.}
\label{fig:System}
\vspace{-.5cm}
\end{figure*}

Although we would ideally need low dimensional features representing the state of the system, our observation consists of high dimensional sensory visual information. This is a complex and highly non-linear function of the system's state. In classical computer vision methods, hand-designed features can be extracted from these high-dimensional observations. 
We recorded videos of task execution by the robot via six high resolution calibrated cameras (with 310x244 pixels and FPS=25). 
We exploit DNNs to extract the low dimensional features (described in Section model). 
Figure\ref{fig:System} shows that the image sequence goes into the CNN (as the feature extractor) and the output flattened layer of the CNN is then concatenated with robot's current state vector; this vector represents system's state and builds the latent vector which we use as a time series for next state prediction. This part of the network is trained independently and has the advantage that the entire model is conditioned on observation at time step $t$. We have tried several CNN architectures including SOTA and well-know CNNs such as VGG19 and ResNet152V2 as well as customised structures; and tried to choose the best model for feature extraction. 
\newline
We performed 60 needle insertion tasks with different tissue configuration where the insertion and exit points are randomly selected and recorded in each experiment with 6 cameras having different poses. 
As such, we collected 360 videos which form our dataset. 
Having 6 different camera views allows studying how the learned D-RLfD model generalises to a new camera view. The camera calibration data represent the pose of the camera frame w.r.t the robot's base frame which is also included in the dataset. 
The future vector extractor DNNs, hence, consists of a CNN whose output is concatenated with robot states, and calibration data (Fig.~\ref{fig:CNN}). 
In our experimental setup Arm 1 has a predefined trajectory and Arm 2 is controlled. 
In our baseline model implementation, our feedforward model (Fig.~\ref{fig:CNN}) uses a few dense layers, namely $\{Dense\_1,\:Dense\_2,\:Dense\_3\}$, taking the concatenated feature vector and generates the predicted control input for the robot. 
The output of the network is the predicted robot's state at time $t+1$, which is a $1\times7$ vector where the first four components are robot's orientation in quaternion and the last three components are robot's Cartesian position. 
\newline
We also implemented SOTA RNN networks suitable for time series (and dynamical behaviour) forecasting as RNN learns the time dependencies of the system's states. Although the acquired accuracy with the CNN could be sufficient for other applications, to increase the precision necessary in safety critical and sensitivity tasks, such as surgical tasks, we have also tested multiple RNN models, such as LSTM and GRU cells.


\section{Experimental Evaluation}
\vspace{-5pt}
In this section, we compare the performance of different D-RLfD architectures including simple feedforward convolutional networks, fully general RNN, GRU, and LSTM models. We have tuned and tested several network architectures. We only include the networks architectures with the best performance in the manuscript where the information about other networks can be found in supplementary materials. 
To measure the performance of the proposed approach, we use a series of metrics, including (i) mean square error (MSE) to illustrate that our approach achieved sufficient amount of accuracy in position and orientation regression; (ii) the maximum position error-- $\mathrm{MaxPE} = \max(e_p)$ where $ e_p(t+1) = \|S_p(t+1)- \hat{S}_p(t+1)\|$, $S_p(t+1)$  and $\hat{S}_p$ are the commanded position and the corresponding value predicted by our D-RLfD at time $t+1$; (iii) average position error, $\mathrm{AvePE} = \frac{1}{n}\sum e_p(t+1)$ where $n$ is the number of samples in each trial; (iv)  the maximum orientation error-- $\mathrm{MaxOE} = \max\:(e_o)$ where $ e_o(t+1)$ is the quaternion distance (see Eq.~\eqref{eq::qua_dis}) between actual commanded orientation and the value predicted by our netweorks; (v) average orientation error, $\mathrm{AveOE} = \frac{1}{n}\sum e_o(t+1)$. 
We have used mean squared error (MSE) which has the Euclidian distance of the actual position commands in the dataset and the one generated by our D-RLfD networks. As for the orientation, we compute the quaternion distance, defined in Eq.~\ref{eq::qua_dis} as the error:
\begin{equation}
\vspace{-0pt}
e_{o} = \frac{1}{T}\sum_{t=1}^{T} (1 - <q_2(t+1), \hat{q}_2(t+1)>^2)
\label{eq::qua_dis}
\vspace{-1pt}
\end{equation}
where $<q1,q2>$ is the inner product of two quaternions. 
$n$ varies across different trials between 170 to 200 samples. In general, we have 65,000 images/samples for the 60 tests. The dataset is split into 80\% and 20\% for training and testing, respectively, where 20\% of training data are for cross validation.
The observation space can be denoted by $O_t^i$ where $i = 1, 2, ..., 6$ is the camera number and $t$ is discrete time. For each time step we have six images and the corresponding robot state $S(t)$. The \textit{Loss} parameter in the tables show the mean absolute error loss function value on the test set for each network

\vspace{-7pt}
\subsection{Base-line Test with feed-forward model}

\label{sec::FFM}
Our baseline implementation contains a CNN, concatenation layer combining latent features vector with robot's state data, and a few dense layers making the prediction of the control commands, as shown in Fig.~\ref{fig:CNN}. 
The desired exit point is marked with a red cross in the videos (as shown in figure \ref{fig:imgsequence}) where the networks needs to output the the desired pose (position $S_p\in \mathbb{R}^3$ and orientation ) of Arm 2 at $t+1$ based on the noisy, high-dimensional pixel data. We have tested various CNN architectures, e.g. AlexNet, VGG19, ResNet and our customised CNN (see Fig.~\ref{fig:CNN}), and the corresponding results are presented in Table \ref{table:1}.
\begin{table}[tb!]
\caption{Evaluation of D-RLfD with CNN architectures}
\centering
 \begin{threeparttable}[b]
\begin{tabular}{ c c c c c c}
 \hline
 \textbf{Model}& MaxPE\tnote{$\dagger$} &  AvgPE\tnote{$\ddagger$} & MaxOE &  AvgOE & Loss\\
 \hline 
 AlexNet   &   0.212    &  1.983  &  2.518  &  1.027  &  0.053\\

 VGG19   &   0.166    &  1.65  &  1.641  &  0.267  &  0.054\\
 
 MobileNet   &   0.173    &  0.825  &  1.077  &  0.389  &  0.049\\
 
 ResNet   & 0.280    &   0.964   &   1.631  &  0.477   & 0.050\\
 
 Our CNN & 0.154 &  0.755  &  1.528  &  0.475  & 0.005 \\
 \hline
 \end{tabular}
 \begin{tablenotes}
     \item[{$\dagger$}] PE in $[mm]$ and OE in $degree$
     \item[{$\ddagger$}] $\mathrm{e}{-3}$
   \end{tablenotes}
 \end{threeparttable}
 \vspace{-12pt}
\label{table:1}
\end{table}
According to the results, the customised model outperforms other tested approaches in terms of MaxPE, AvePE, MaxOE and AveOE (position values are in \textit{mm} for $xyz$ translation and orientation values are in \textit{degree} for euler angles). In this specific application the maximum error value plays a vital role; since larger bound on the maximum error indicates the risk of damaging the tissue under operation and our CNN yields very small MaxPE and MaxOE values indicating an acceptable margin of needle's tracking error during a suturing task. 
\vspace{-7pt}
\subsection{One-Camera as Test set with feedforward model}

\label{sec:FF}

\begin{table}[tb!]
\caption{Evaluation of the D-RLfD with CNN architecture: test-set includes only one camera data unseen in training.}
\centering
 \begin{threeparttable}[b]
\begin{tabular}{ c c c c c c}
 \hline
 \textbf{Test Camera}& MaxPE\tnote{$\dagger$} &  AvgPE\tnote{$\ddagger$} & MaxOE\tnote{$\dagger$} &  AvgOE\tnote{$\dagger$} & Loss \\
 \hline 
 Camera1   &   0.30    &  0.61  &  14.90  &  2.40  & 0.021\\

 Camera2   &   0.26    &  0.79  &  6.66  &  2.33 & 0.014 \\
 
 Camera3   &   0.19    &   0.44   &   8.42  &  2.17  & 0.012 \\
 
 Camera4   &   0.27    &   0.89   &   11.68  &  2.25  & 0.019 \\
 
 Camera5   &  1.41    &   4.24   &   14.96  &  2.51  & 0.018 \\
 
 Camera6   &   1.62    &   2.89    &   5.90  &  2.56 &  0.023 \\
 \hline
 \end{tabular}
 \begin{tablenotes}
     \item[{$\dagger$}] PE in $[mm]$ and OE in $degree$
     \item[{$\dagger$}] $\mathrm{e}{-2}$
   \end{tablenotes}
 \end{threeparttable}
 \vspace{-13pt}
\label{table:2}
\end{table}
In Section~\ref{sec::FFM}, we used all the 6 camera views to form the training and test set. 
However, this does not illustrates whether the learned model performs well if the camera is located at a new pose. 
We are also interested in understanding how a model trained on some camera views generalises to a new camera view unseen during training. So, we considered data corresponding to 5 camera views as the training set and the data corresponding to the 6th camera view as the test set. Table \ref{table:2} presents the results where one Camera $i$ is excluded in the training and used only for testing. We only use the customised CNN shown in Fig.~\ref{fig:CNN} in this section. Hence, the error values in Table \ref{table:2} must be compared to those in the last rows of Table \ref{table:1} in which all six cameras are included in the training set. As table \ref{table:2} shows only the position error values for camera5 and camera6 are relatively higher but considering the $10^{-2}$ order still in an acceptable range; and it is due to the relative distance of the third stereo pair to the other two.
\vspace{-4pt}
\begin{table}[b!]
\centering
\begin{threeparttable}[b]
\caption{Evaluation of RCNs including fully general RNN, GRU and LSTM.}
\begin{tabular}{ c c c c c c c}
 \hline
 \textbf{Model} &   MaxPE\tnote{$\dagger$} &  AvgPE\tnote{$\dagger$} & MaxOE\tnote{$\dagger$} &  AvgOE\tnote{$\dagger$}  & Loss\\
 \hline 
 \hline
 
 Feedforward     &  0.154 &  0.755  &  1.528  &  0.475  & 0.005 \\ 
 
 RNN     &   0.068 &  0.153  &  0.704  &  0.173  & 0.0097\\
 
 GRU     &   0.029 &  0.188  &  0.370  &  0.119 &  0.0051 \\
 
 LSTM    &   0.035 &  0.134  &  0.338  &  0.106  & 0.0048 \\
 
 \hline
\end{tabular}
   \begin{tablenotes}
     \item[{$\dagger$}] PE in $[mm]$ and OE in $degree$
     \item[{$\dagger$}]  $\mathrm{e}{-4}$
   \end{tablenotes}
 \end{threeparttable}
 \vspace{-12pt}
\label{table:3}
\end{table}
\vspace{-2pt}

\subsection{Base-line Test with RCNs}
To further improve the performance D-RLfD model to generate a time series of control commands, we used the CNN with the best performance in the Section~\ref{sec::FFM} (which generates the latent vector) and build several recurrent networks. RNNs enable the model to use both the observation and dynamic of the system to generate the control commands. Table 3 presents the results for RNN, GRU, LSTM and simple feed-forward model. All the recurrent networks yield improved precision and accuracy of the position and orientation compared to the feed-forward model where the results indicate LSTM has the lowest error values.
\vspace{-4pt}
\begin{table}[tb!]
\caption{Results of the feed-forward D-RLfD with augmented calibration data corresponding with the results presented in Table~\ref{table:2}.}
\centering
 \vspace{-7pt}
 \begin{threeparttable}[b]
\begin{tabular}{ c c c c c c}
 \hline
 \textbf{Test Camera} & MaxPE\tnote{$\dagger$} &  AvgPE\tnote{$\dagger$} & MaxOE\tnote{$\dagger$} &  AvgOE\tnote{$\dagger$} & Loss \\
 \hline 
 Camera1   &   \textbf{0.19}    &  0.70  &  \textbf{7.28}  &  \textbf{2.05}  & \textbf{0.019} \\

 Camera3   &   0.31    &   0.58   &   \textbf{4.93}  &  4.60  & \textbf{0.010} \\
 
 Camera5   &  \textbf{1.07}    &   \textbf{0.78}   &   \textbf{8.53}  &  \textbf{1.50}  & \textbf{0.015}\\
 
 \hline
 \end{tabular}
 \begin{tablenotes}
     \item[{$\dagger$}] $\mathrm{e}{-3}$
   \end{tablenotes}
 \end{threeparttable}
 \vspace{-3pt}
\label{table:4}
\end{table}
\vspace{-0cm}
\subsection{Calibration data Augmentation}
\vspace{-5pt}
Thus far, we have used the image data and robot's current state, i.e. kinematic data, as the input to our D-RLfD models. However, calibration data carries some relevant information of cameras position/orientation w.r.t robot's base frame which may improve the performance of our models. We, thus, refined our baseline feed-forward model so that the calibration data is also concatenated with the latent feature vector (figure \ref{fig:CNN}). Here, for the proof of concept we only present the results obtained by refined feed-forward model (whose basic implementation results are presented in Table~\ref{table:2}) with augmented calibration data of Camera 1, 3 and 5. Table \ref{table:4} presents the results of feed-forward D-RLfD model with augmented calibration data which indicates improved accuracy of control command prediction in terms of \textit{Loss} and some of the errors (presented with bold face). However, some of the error values became worse (probably) because we concatenated the calibration data with the feature vector and network does not have enough layer after that to learn the correlation between the calibration data and feature vector. 
%
\section{Conclusion}
\label{sec:conclusion}
This paper presents a dataset of inserting a circular needle with a da Vinci Research Kit (DVRK) into a soft tissue to be passed through the desired given exit point.  The dataset is suitable for data-driven approaches and consists of 60 successful needle insertion trials. The videos of each trial along with the corresponding robot's data, calibration parameters and commanded robot's inputs are included in the dataset. 
We also presented some baseline Deep-Robot Learning from Demonstration (D-RLfD) networks. Our results show RCNs outperform simple feed-forward NNs (as expected), where augmenting the camera calibration data with the feature vector (outputs from the CNNs) improves some of the error metrics whereas other error metrics get slightly worse. Moreover, the baseline feed-forward networks trained on 5 camera view as train sets can successfully generalise to the unseen camera view captured by camera 6. 
\newline
In future works, we will study more sophisticated D-RLfD architecture to improve the performance of control command inference. We will also consider different architectures with enough layers to capture the correlation between the feature vector and calibration data and study a normalised error metric as Euclidean and quaternion distances may be expressed in different scales. Moreover, we will consider an extensive study on generalising to an unseen camera view. 


\bibliographystyle{IEEEtran}
\bibliography{example} 

\end{document}